\documentclass[letterpaper, 10 pt, conference]{ieeeconf}  

\IEEEoverridecommandlockouts                              

\usepackage{amsmath,amssymb,amsfonts}
\usepackage{algorithmic}
\usepackage{graphicx}
\usepackage{textcomp}
\usepackage{xcolor}
\let\labelindent\relax
\usepackage{enumitem}
\usepackage[utf8]{inputenc} 
\usepackage[T1]{fontenc}
\usepackage{url}
\usepackage[hidelinks]{hyperref}

\title{\LARGE \bf
Towards Understanding Ambiguity Resolution in Multimodal Inference of Meaning
}

\author{Yufei Wang$^1$ \and Adriana Kovashka$^1$ \and Loretta Fern\'andez$^2$ \and Marc N. Coutanche$^3$  \and Seth Wiener$^4$%
\thanks{$^{1}$Department of Computer Science, University of Pittsburgh, Pittsburgh, PA, USA. {\tt\small \{yuw384, kovashka\}@pitt.edu}}%
\thanks{$^{2}$Department of Teaching, Learning, and Leading, University of Pittsburgh, Pittsburgh, PA, USA. {\tt\small lof7@pitt.edu}}%
\thanks{$^{3}$Department of Psychology and Learning Research and Development Center, University of Pittsburgh, Pittsburgh, PA, USA. {\tt\small marc.coutanche@pitt.edu}}%
\thanks{$^{4}$Department of Languages, Cultures, and Applied Linguistics, Carnegie Mellon University, Pittsburgh, PA, USA. {\tt\small sethw1@cmu.edu}}%
}

\begin{document}

\maketitle
\thispagestyle{empty}
\pagestyle{empty}

\begin{abstract}

We investigate a new setting for foreign language learning, where learners infer the meaning of unfamiliar words in a multimodal context of a sentence describing a paired image. We conduct studies with human participants using different image-text pairs. We analyze the features of the data (i.e., images and texts) that make it easier for participants to infer the meaning of a masked or unfamiliar word, and what language backgrounds of the participants correlate with success. We find only some intuitive features have strong correlations with participant performance, prompting the need for further investigating of predictive features for success in these tasks. We also analyze the ability of AI systems to reason about participant performance, and discover promising future directions for improving this reasoning ability.

\end{abstract}

\section{INTRODUCTION}
\label{sec:intro}

More than a billion people around the world learn English as a second language \cite{BritishCouncil2021English}. In Europe, 96\% of secondary students study a foreign language; China has more than 300 million English learners \cite{Eurobarometer2024}. In the United States, 20\% of the population, or roughly 68 million people, able to speak a language other than English \cite{censuswebsite}. 
The rise of language learning apps, like Duolingo with its 74 million monthly users \cite{Duolingo2023}, highlights the growing demand for multilingualism in the workforce. The question is how do we train the next generation to become effective multilinguals? 

An immersive and interactive environment is considered ideal for learning foreign languages \cite{shrum2015teacher}. Immersive learning contextualizes the meaning of words, often through imagery, and grounds language in a real-world setting. When learners interact with language in real-world scenarios, they are compelled to decode meaning, predict outcomes, and adapt their understanding based on immediate feedback from the environment. This contrasts sharply with passive learning models that rely heavily on rote memorization, where context is often stripped away, leaving learners with fragmented knowledge that is harder to apply. Further, the active engagement required in real-world settings can build confidence and intrinsic motivation, as learners gradually experience success in navigating unfamiliar linguistic and cultural terrains. 

Imagine going to a farmers market during a study abroad and listening to the vendor describe the items for sale. The vendor might point to some objects, but the pointing may be ambiguous, as multiple items could be the target of the pointing. The descriptors could similarly refer to multiple objects, given the listener's imperfect knowledge of the language. How can the learner make sense of what is being said, ground (associate) some of the unfamiliar spoken concepts to their visual manifestation (i.e., their correct meaning), and thus acquire new concepts? 

In language education, the ability to navigate new unfamiliar or uncertain situations without experiencing frustration is termed ambiguity tolerance \cite{chu2015relationships}. 
In this paper, ``ambiguity'' denotes the uncertainty of meaning when a learner faces an unfamiliar term, i.e. the multiple possibilities (guesses) of what this unfamiliar term refers to given the incomplete (for the learner) information. Studies show that tolerance for ambiguity is strongly related to success in learning a foreign language \cite{furnham2013tolerance,li2016ambiguity,swaab2003understanding,yu2022review,zarfsaz2014silence,bardi2009openness,wang2020exploring}. 
Helping students develop higher levels of tolerance to ambiguity can help them develop a growth mindset about learning the language \cite{suman2023flmindset} that improves motivation, proficiency, and perseverance. 

While the above work shows the critical role of ambiguity, there is a gap in understanding how second language learners deal with ambiguity in a \emph{multimodal} context, including the optimal level of ambiguity for successful second language learning from specific \emph{data} (examples from which the learner acquires the language). We aim to answer the questions: What is it about a multimodal example (e.g. image paired with text containing unfamiliar terms) that makes it difficult for the learner to correctly infer the meaning of those terms? How well can artificial intelligence support human learning by reasoning about which examples would be challenging, which would then enable the AI system to curate a progressively more challenging sequence of examples?

We focus on a setting where a visual environment (captured in a photo) is accompanied by text or speech, and this text/speech mentions unfamiliar words in a language a participant is learning. 
An example could be a task where a word is replaced by a blank to simulate/ensure a setting where the learner is unfamiliar with this word and has to infer its meaning from the context. This task assesses comprehension, recall, and understanding of specific content, such as vocabulary. This type of exercise is often used in education to reinforce learning or test knowledge in a focused and structured way \cite{nation2013vocabulary}, but there is a lack of analysis of the features of the image or text. 

\begin{figure}[t]
    \centering
    \includegraphics[width=1\linewidth]{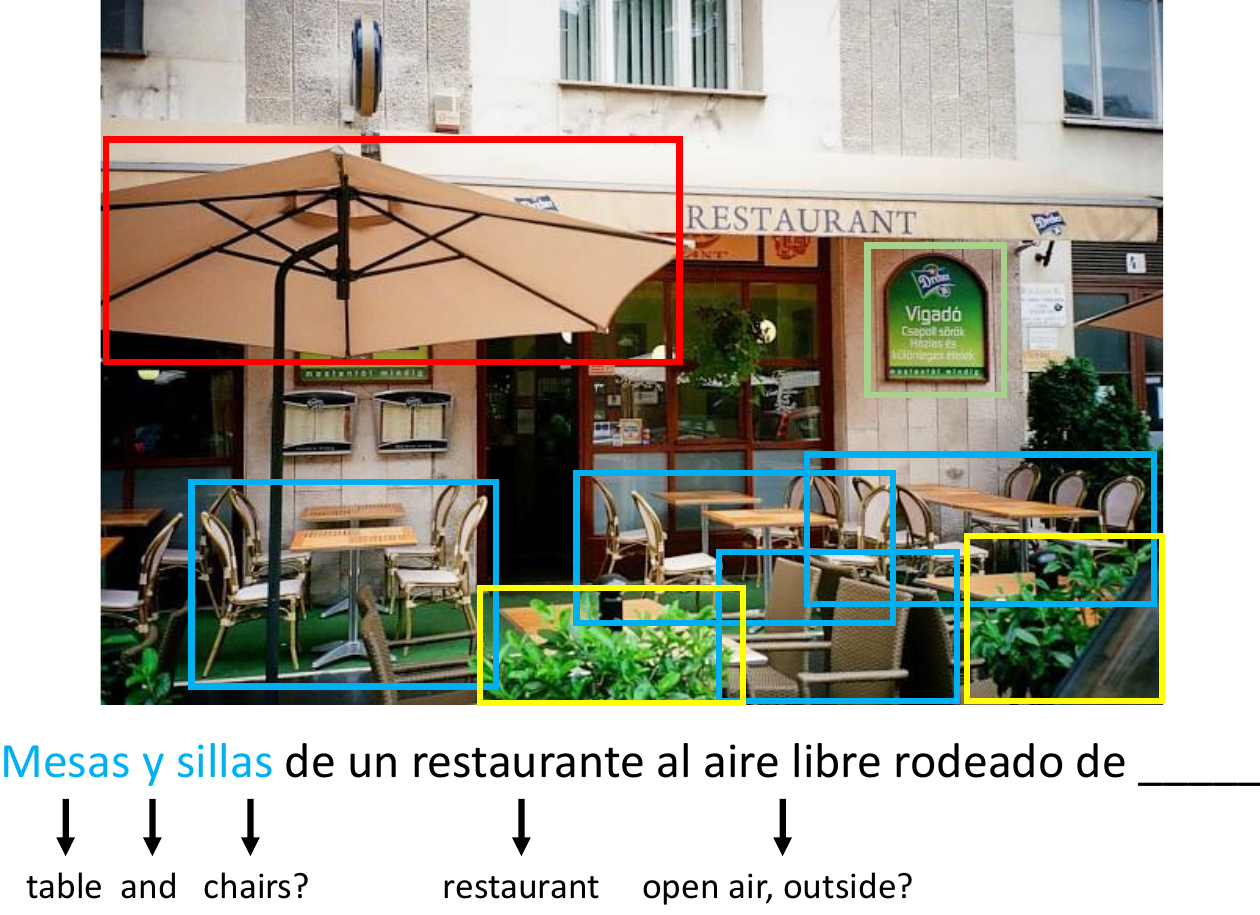}
    \caption{An example meaning inference task.} 
    \label{fig:example}
\end{figure}


Consider a learner of Spanish who is shown a picture Fig.~\ref{fig:example} with the description shown at the bottom and asked to 
fill in the blank. 
This resembles a real-life situation where the learner encounters an unfamiliar word and must resolve it to complete a task, e.g. retrieve an object or identify a meeting location while in a foreign country.
If the learner knows the meaning of ``mesas'' (tables) and ``y'' (and), they might infer that ``sillas'' probably refer to chairs. Thus, they can rule out the regions indicated in blue; these are already referred to by ``mesas y sillas''. The learner might infer the meaning of ``aire libre'' based on similar English words; thus ``de un restaurante al aire libre'' does not mention any further objects, but rather the setting. 
This leaves the possibility that the blank refers to any of the plants (in yellow), umbrella (in red), or sign (in green). Factors that contribute to the difficulty of the task can be found both in the sentence (e.g., the sentence length) and in the image (e.g., the number of objects). 
We quantitatively investigate the diverse factors that contribute to the difficulty of this task, examining the image, text, and participant's language background.

Further, we envision a system which can curate a sequence of examples of progressively greater difficulty, to make learning accessible yet appropriately challenging. This is akin to mediation and dynamic assessment: guided support in which learners receive real-time, adaptive support to overcome challenges and develop their language skills and problem-solving \cite{Lantolf2014sociocultural}. 
The process relies on the concept of Zone of Proximal Development (ZPD), defined as the gap between what an individual can achieve alone, and what
they can achieve when aided \cite{vygotskiui1997collected}. Studies in desirable difficulties have also shown that learning conditions that feel hard in the moment can lead to better long-term retention and transfer \cite{bjork2011making}. Leveraging an AI system to implement such mediation has the potential to enable timely and adaptive scaffolding based on the learner’s responses. 

Thus, we test the potential for an AI system to provide this dynamic curation, by investigating whether an AI system can predict what examples would be easier or harder, based on features of the data (image and text) and participant (language background) and examples of stategies that a learner might use to infer meaning.
\section{BACKGROUND}
\label{sec:background}

Our work aims to bridge theoretical insights with practical applications by linking the tasks used, such as multimodal fill-in-the-blank tasks, to current evidence-based pedagogy. These tasks align with the principles of experiential language learning, which emphasize authentic, context-rich, and interactive experiences for learners \cite{Kolb1984Experiential,Ellis2005Principles}. Research has demonstrated that multimodal and interactive tasks enhance learners' ability to negotiate meaning, improve retention, and develop deeper cognitive connections to language use \cite{Vandergrift2012Metacognition,Swain2013Languaging}. Moreover, integrating such tasks into language instruction resonates with evidence-based pedagogical practices, such as scaffolding \cite{Wood1976Scaffolding}, multimodal input \cite{Mayer2024multimedia}, and gamification \cite{Plass2015Gamebased, gee2013games}, which have been shown to increase engagement, reduce cognitive overload, and support long-term retention. The act of mapping novel words to one of several meanings can itself support the integration of new words into human memory networks \cite{coutanche2014fast}. 

\textbf{Foreign language learning with multimodal aids.}
Language is only one aspect of meaning-making, as meaning is also constructed through non-verbal modes such as visual, auditory, and multimedia elements \cite{kress2011discourse}. Communication is an inherently multimodal process shaped by cultural and social context   \cite{fernandezmultiliteracies}. Each mode of communication possesses its unique affordances and constraints \cite{kress2011discourse}; the combination of modes can help students learn better. 
Research has shown that multimodal learning, where visual, auditory, and kinesthetic inputs are integrated, improves memory consolidation and retrieval \cite{Paivio1986Dualcoding}. Additionally, multimodal settings often incorporate authentic and dynamic interactions, which can enhance the intrinsic motivation by making the material more engaging and contextually relevant \cite{Dornyei2011Motivation}. 
Some work examined learning English from movies \cite{kabooha2016using,albiladi2018learning,goctu2017using,csomay2012yes,roslim2021exploring}, and reported learners find it enjoyable and effective. 
Works in language development  analyzed how children interact with multimodal information to learn the meaning of words \cite{Medina_2011_Human_Simulation_Paradigm, Yurovsky_2013_Human_Simulation_Paradigm} and explored the problem of referential uncertainty \cite{referential_uncertainty_Yu_2021} and the strategy of mutual exclusivity \cite{mutual_exclusivity_Markman_1988}. We build on this foundation to explore ambiguity resolution  specifically in foreign language learning under a multimodal setting.

\textbf{Foreign language learners dealing with ambiguity.}
Exposure to unfamiliar, complex, and ambiguous situations, such as in different cultures and languages, can lead to confusion. Yet, ambiguity is inherent in all languages and a learner will inevitably encounter it. Recent evidence-based pedagogy suggests ambiguity is valuable and instructors should embrace it in the classroom \cite{richardson2016toward, kubanyiova2020language, dewaele2013link}.
Numerous works document a relationship between ambiguity tolerance and motivation to actively participate in classroom activities \cite{yu2022review,zarfsaz2014silence} or tolerance of academic challenges
\cite{bardi2009openness,wang2020exploring}. 
We focus on the aspects of the ambiguous \emph{examples} to understand what makes particular learning \emph{materials} appropriate.

\textbf{Using AI to understand how children learn meaning.}
Our work bears some resemblance to work which uses artificial intelligence tools to understand human learning, specifically children learning nouns and verbs, including using video data \cite{yurovsky2013statistical,leung2021parents,zhou2025quantifying}. For example, \cite{zhou2025quantifying} explore correlations between visual and textual features from an AI model, to the age at which a child learns specific verbs in their native language. Conversely, there is some work which uses data collected by children (e.g. via wearable cameras) to train artificial intelligence models \cite{sheybani2024curriculum,bambach2018toddler,vong2024grounded}. None of this work applies to foreign language learning: using AI to curate examples for foreign language learning. 

\begin{figure*}[t]
    \includegraphics[width=1\textwidth]{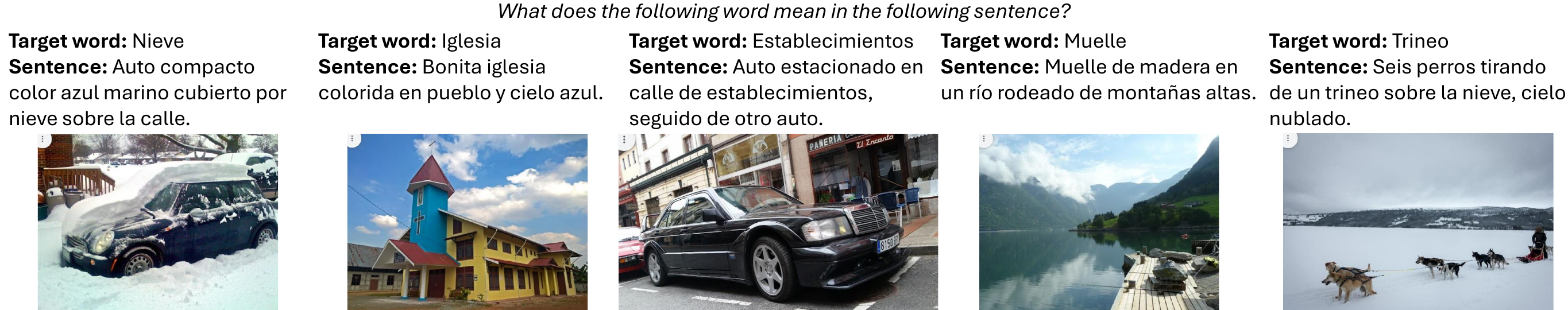}
    \caption{Five image-text pairs from our preliminary word meaning inference experiment.}
    \label{fig:wordguess}
\end{figure*}

\textbf{Vision-language datasets.}
Resources leveraged for training artificial intelligence (AI) systems offer an excellent source of data through which we can study inferring meaning in foreign language learning. 
VL data have shown great promise to advance computer vision tasks \cite{Radford2021LearningTV,wslimageseccv2018,Jia2021ScalingUV} by learning visual and text representations from 
co-occurrence relationships between the two modalities, where the similarity is maximized between an image and its accompanying caption.
We rely on descriptive captions, such as in the COCO dataset \cite{Lin2014MicrosoftCC}, follow the image closely and mention most relevant details, e.g. ``there is a man cooking hamburgers at a grill''. 
There are also captions that are written more naturally and loosely to complement an image, e.g. text that social media users write when uploading their photos, e.g. in RedCaps \cite{desai2021redcaps,Ordonez2011Im2TextDI,Changpinyo2021Conceptual1P}. 
Most works focus on English, but some go beyond, e.g. in XM3600
\cite{thapliyal2022crossmodal,chen2022pali,pfeiffer2022xgqa,jain2021mural,kim2020mule,burns2020learning,liu2021visually}.
However, vision-language data has not been used to analyze and improve foreign language learning.
\section{PARTICIPANT STUDIES}
\label{sec:studies}

We study factors affecting the difficulty in inferring the meaning of a word in an online multimodal setting. 
Participants are asked to (1) recognize the visual context presented in an image, (2) read a corresponding sentence in a foreign language, and (3) infer the meaning of a specific word based on the visual context and the sentence provided. Responses are written in English, and we do not provide feedback.

We conducted two experiments. The first featured a randomly chosen set of 50 image-text pairs. The second featured 10 image-text pairs manually curated to not be \emph{too} easy (too few objects or words too similar to English) or \emph{too} hard, but to show a slight progression from easier to harder. We aimed for consistent reference to the same target word across languages. 
We used XM3600 data \cite{thapliyal2022crossmodal}. 

The first study used the task illustrated in Fig.~\ref{fig:example}. We created a form in Spanish with 50 image-text pairs, with a single noun in each sentence replaced with a blank. 
Eight participants completed the study; seven had very limited Spanish knowledge, and one had intermediate knowledge.
We manually compared the answer provided to the original word in the blank, and considered the response correct if it was an exact match or closely related (e.g. flower-plant). 

In the second study, we used a smaller set of images with captions (five examples shown in Fig.~\ref{fig:wordguess}), but created the tasks in five different languages (Spanish, French, German, Korean and Turkish), and also collected information about the languages the participants were familiar with (indicated by the participant using a scale of 1-5 for proficiency). 
We clarified that the target word to guess is always a noun. To provide a reference for the pronunciation of words, we provided Korean sentences in both original and romanized formats. We asked participants to attempt the task in the language they had the least background in (but still with some familiarity or somewhat lexically similar with their native language)
first, and the language with most background in last. We also asked them to identify the words in the sentence whose meanings they could determine and to explain the process they used to infer those meanings. Participants were instructed \emph{not} to use Google Translate or similar services. We obtained about 10 responses per language. We calculated accuracy in a generous way (e.g. ``restaurant'' and ``establishment'' were both correct, ``tree'' and ``forest'' were both correct). Among our participants, 
5 had some familiarity with Chinese, 4 with Korean, 3 with Japanese, 4 with Turkish, 2 with Persian, 5 with German, 5 with Spanish or Italian, 2 with French, and all were advanced or fluent in English. On average participants were somewhat familiar with 3 languages (excluding English) from 2 language groups (grouping languages by European, West Asian and East Asian). We excluded native speakers. 
\section{ANALYSIS AND DISCUSSION}
\label{sec:analysis}

\subsection{Data metrics}
\label{sec:data_metrics}

\begin{table*}
\caption{Correlations (P for Pearson, and S for Spearman) between data features and guessing performance. Significant values starred (two stars for strong and one star for weak significance; see text). Manual and Auto refer to the manner of computing the features, and 25/50 refers to the number of samples for which features were computed.}
\centering
    \begin{tabular}{|c|l|l|l|l|l|l|}
    \hline
    Features & Manual (P-25) & Manual (S-25) & Auto (P-25) & Auto (S-25) & Auto (P-50) & Auto (S-50) \\ \hline \hline
    Number of objects & -0.4012** & -0.3106 &-0.2193 &-0.1444 &-0.1352 &-0.0659\\ \hline
    Size of target & 0.0338 & -0.0336 &-0.2228 &-0.2808 &-0.1433 &-0.1435\\ \hline
    Location of target & -0.1248 & -0.0766 &0.0661 &0.0151 &0.0473 &0.0304\\ \hline 
    \hline
    Object touched? & -0.0513 & -0.0766 &0.0453 &0.0113 &-0.1983 &-0.0310 \\ \hline \hline
    Sentence length & -0.4758** & -0.4849** &-0.4758** &-0.4849** &-0.1981 &-0.1695\\ \hline
    Distance from start & -0.3117 & -0.2576 &-0.3117 &-0.2576 &-0.1037 & -0.0970 \\ \hline
    Distance from end & -0.0535 & -0.0625 &-0.0535 &-0.0625 &-0.0468 &-0.0761\\ \hline 
    Fraction of nouns & - & - &0.1820 &0.0357 &0.2666* & 0.2025\\ \hline \hline
    CLIP (English) & - & - & 0.0721 & 0.0219 & 0.0794 &0.0733\\ \hline
    CLIP (Spanish) & - & - & 0.1646 & 0.1534 & 0.0348 &0.0152\\ \hline    
    \end{tabular}
    \label{tab:corr_data}
\end{table*}


We consider the following \emph{text factors}, which we expect may correlate with overall success of word meaning inference on a particular example (measured as the fraction of participants who guessed this word's meaning correctly): 
(1) sentence length (number of words; longer sentences may be harder to parse); (2) proximity of the target word to the sentence start (more prominent objects may be described first and be easier to guess) or end; 
(3) the fraction of nouns in the sentence, out of all words.
It may be that the more nouns there are, the easier for a participant to recognize their meaning and/or recognize them in the image, and rule them out as possibilities for the target word's meaning. The opposite may also hold, where fewer nouns can lead to concise sentences that make the task easier. There is limited prior work to decisively support either hypothesis.

For \emph{image factors}, we consider:
(1) the number of objects present (more objects would make the task harder);
(2) the size and location of the objects (the larger the object, the more likely to be described, probably early in the sentence, making it easier to guess \cite{hwang2010accounting}); 
(3) whether persons in the image are handling an object or not (objects that are handled may be more likely to be mentioned and vice versa); 
(4) scores for each image-text pair from CLIP \cite{radford2021learning}, a pretrained AI model which learns image and text representations in such a way as to maximize the dot product (similarity) between images and text that appear together on the web (we expect easier guessing in visual-text pairs with high CLIP score).

\subsection{First study analysis: Data metrics}

We begin with some general observations: 
\begin{enumerate}[nolistsep,noitemsep]
\item The average number of correct guesses per image was 4.94 (out of 8), and std. deviation was 2.02, confirming individual samples have different difficulty. 
\item It was easier to guess in simpler images with four or fewer unique objects; P(easy$|$simple) = 0.67 vs P(easy$|$complex) = 0.52, where ``easy'' is defined as samples with over 5 correct guesses (out of 8).
\item Blanks in shorter sentences were easier to solve (success rate 0.64 in short sentences) than in longer sentences (success rate 0.43 in longer sentences); short sentences were ones less than 12 words long. 
\item Image-text similarities from CLIP \cite{radford2021learning} were 1 point higher for successfully resolved samples. 
\end{enumerate}

Table \ref{tab:corr_data} shows the correlations we obtained from the data features (Sec.~\ref{sec:data_metrics}) and performance in guessing. We use two sets of feature values. The first uses manual annotations by one of the authors. The second uses an AI system to compute the values: InternVL \cite{chen2024internvl} for image/object features, CLIP \cite{radford2021learning} for CLIP features, GPT-4o \cite{openai2024gpt4ocard} for fraction of nouns, and manual code computation/counting for the rest of text features. We only manually labeled the first 25 image-text pairs, so we show correlations based on automatic features of the first 25, and the full set of 50 image-text pairs. We show both Pearson and Spearman correlations. We denote significance at $p<0.05$ with two stars, and at $p<0.10$ with one star. We observed that very few features of the data were found to be significantly correlated with guessing performance. Number of objects was significantly negatively correlated with success in the task (using Pearson correlation)---the more objects in the image, the harder the task. 
Sentence length was also significantly negatively correlated, with both correlation metrics---the longer the sentence, the harder the task. 
Fraction of nouns was positively correlated with success under Auto (P-50); we hypothesize the more objects/nouns in the image, the easier it was to match words to visual content and rule out possible matches for the target word.
However, no other features were significantly correlated with success in the guessing task.
This is an interesting observation demonstrating that predicting guessing success requires further research. 

\subsection{Participant background metrics}

We tracked the following features for our participants: 
\begin{itemize}
    \item Proficiency in target language, 5 (fluent) to 1 (beginner).
    \item Sum of proficiencies in related languages. Due to the limited number of languages we study, we opt for a coarse grouping: for European languages, Romance or not; for Asian ones, West or East.\footnote{A more sophisticated grouping could use the World Loanword Database \cite{Haspelmath2009Loanwords} and the Automated (phonetic) Similarity Judgment Program    \cite{Holman2011LanguageFamilies}.} For Spanish and French, the group of related languages was Spanish, French and Italian. For Korean, it was Korean, Chinese, Japanese and Shanghainese. For Turkish, it was Arabic and Hindi. For German, it was only German. 
    \item Number of languages spoken, (from self-reported list).
\end{itemize}

\subsection{Second study analysis: Participant metrics}

We observed the following:
\begin{enumerate}[nolistsep,noitemsep]
    \item The task in Spanish had the highest average accuracy (Acc) (across participants) of 7.1 out of 10 questions with standard deviation (SD) 1.8 and median (Med) 7. This was followed by Korean (Acc 6.6, SD 2.3, Med 7), German (Acc 6.4, SD 2.1, Med 7), French (Acc 6.2, SD 1.5, Med 6) and Turkish (Acc 6.2, SD 1.9, Med 6). 
    \item Familiarity with a language was not sufficient nor necessary for above-average performance. While in general those familiar with a language did better, participants who were not familiar also in some cases did very well.  
    \item Participants varied in their ability to recognize cognates to English words, despite shared English proficiency. 
\end{enumerate}

\begin{table}[]
 \caption{Correlations (P=Pearson, S=Spearman) between features and guessing accuracy (significant values starred).}
    \centering
    \begin{tabular}{|c||l|l|l|l|l|}
    \hline
    Feature & Spa & Fre & Ger & Tur & Kor \\ 
    \hline \hline
    Target (P) & {0.537*} & 0.368 & {0.647**} & {0.806*} & {0.849**} \\ 
    \hline
    Target (S) & 0.398 & 0.357 & 0.520 & {0.850}** & {0.888}** \\ 
    \hline \hline
    Related (P) & 0.283 & 0.325 & {0.647**} & 0.712 & {0.612*} \\ 
    \hline
    Related (S) & 0.278 & 0.299 & 0.520 & {0.764*} & {0.684**} \\ 
    \hline \hline
    Number (P) & 0.460 & 0.373 & 0.346 & 0.057 & 0.331 \\ 
    \hline   
    Number (S) & 0.450 & 0.377 & 0.298 & 0.242 & 0.344 \\ 
    \hline
    \end{tabular}
    \label{tab:corr_lang_background}
\end{table}

Table \ref{tab:corr_lang_background} shows correlations between the three participant metrics, and guessing success. Surprisingly, for several languages, background in related languages is not significantly correlated with performance. The most significant values are in Turkish and Korean, which is intuitive since these are languages where specialized knowledge is needed beyond the shared English proficiency, while for European languages, familiarity with English is more helpful. In Spanish and French, number of languages spoken had high, but nonetheless not significant, correlation with performance.

\subsection{Strategies used: Second study}

Below, we list the strategies participants reported using, in order to infer word meaning.
One is the exclusion principle: if participants are familiar with an object term that is not the target word, they can map this to an object in the image, thus reducing the number of possibilities for the object in the image the target word could refer to \cite{coutanche2015rapid,halberda}.

A second strategy is using grammar, e.g. determiners to identify nouns, or prepositions to narrow down the possible objects that follow a specific spatial relationship---for example, only a limited set of objects in an image are in an ``on top of'' relationship. 
A related strategy is to leverage the structure of a particular language, e.g. knowing that locations are listed last, or that adjectives follow nouns. 

A third strategy involves analyzing the placement and size of objects in the image. More prominent (central or bigger) objects are likely to be mentioned, perhaps exclusively or earlier in the sentence \cite{hwang2010accounting}. Conversely, objects in the background may not be mentioned. This narrows down the possibilities for inferring word meaning. 

A fourth strategy relies on word similarity, potentially phonetically and not lexically. For example, one participant identified a relationship between ``iglesia'' in Spanish and ``kilise'' in Turkish, both meaning church. Some similar-sounding words may be false cognates, but these may  be ruled out based on context.

The least commonly used strategy was using grammar. Exclusion principle was sometimes used, but in some cases, incorrectly (not all possibilities were considered and ruled out). The following are examples of strategies that only one participant used for a particular image-text pair:
\begin{itemize}[nolistsep,noitemsep]
    \item \underline{Grammar:} ```vor einem' sounds like 'above' or 'in front of'; I identified words that may refer to the car and parking in the first half of the sentence; I would guess the target word is a location'' (note participant reported no background in German) 
    \item \underline{Grammar:} ``Based on the words that I recognized, I think the sentence is referring to two trains stopping in the station. As ``Bahnhof" appears after ``im", it should be the location.''
    \item \underline{Grammar:} 
    ``I guess seis perros tirando means six dogs; and un means that there's only one of the target word in the picture; it could be the driver or the sleigh.''
    \item \underline{Similar words:} ``The focus of image is a mini Cooper and the snow. The mini Cooper is mentioned in the sentence as another word, so my guess was snow'' (note the speaker had no background in Korean, but mapped ``minikupeo'' to ``mini Cooper'')
    \item \underline{Similar words, exclusion:} ``There is compact car and snow on the photo. Auto compacto should be compact car so Nieve should be snow but I am not sure.''
    \item \underline{Exclusion:} ``I recognize rio as water, and guess the other half of the sentence is talking about the trees''
    \item \underline{Exclusion:} ``I identified mountain and on the water, and guess the target word should be the harbour''
    \item \underline{Exclusion:} ``the first half is describing the train so i assumed since its referring to it ``in a [blank]' it would be that it is ``in'' a station''
\end{itemize} 

Future research could explore how to equip learners with better command of these strategies, e.g. via an AI system to select or generate examples whose resolution requires using a particular strategy. This motivates the next exploration.

\section{ABILITY OF AN AI SYSTEM TO PREDICT PERFORMANCE}
\label{sec:ai}

\begin{table*}
\caption{AI system ability to guess participant performance. Features: (B)ackground / (R)ecognized word/ (S)trategy / (S)trategy (S)ummary. Images: + represents that the model has the original image file as input while - indicates no information regarding the image. ``Text'' stands for having textual descriptions of the image as part of the input. The highest performance in each column per group (top, middle, bottom, depending on use of image) is shown in bold, while the second best is italicized.}
\centering
\begin{tabular}{|c|c|c||c|c|c|c|c|c||c|}
\hline
Model	& Feature & Image	& Spanish	& French	& German	& Korean/Normal	& Korean/Roman	& Turkish	& Average \\
\hline \hline
InternVL2.5	& B	& + & 0.517	& 0.491	& 0.509	& 0.480	& 0.310	& \textbf{0.550}	& 0.476 \\
\hline
InternVL2.5	& B+R & + & 0.558	& 0.482	& 0.509	& 0.450	& 0.410	& 0.383	& 0.465 \\
\hline
InternVL2.5	& B+S & +	& 0.558	& 0.473	& 0.564	& \textit{0.570}	& \textbf{0.490}	& \textit{0.433}	& 0.515 \\
\hline
InternVL2.5	& B+R+S	& + & \textbf{0.642}	& \textit{0.500}	& \textbf{0.600}	& \textit{0.570}	& \textit{0.480}	& 0.367	& \textit{0.526} \\
\hline
InternVL2.5	& B+SS & +	& \textit{0.625}	& \textbf{0.591}	& \textit{0.573}	& \textbf{0.600}	& 0.470	& \textbf{0.550}	& \textbf{0.568} \\
\hline \hline
InternLM2.5	& B & -	& 0.475	& 0.482	& 0.445	& 0.420	& 0.290	& 0.300	& 0.402 \\
\hline
InternLM2.5	& B+R & -	& 0.508	& 0.409	& 0.436	& 0.390	& 0.370	& 0.283	& 0.400 \\
\hline
InternLM2.5	& B+S & -	& \textbf{0.650}	& \textit{0.518}	& \textbf{0.564}	& \textbf{0.570}	& \textbf{0.510}	& \textit{0.383}	& \textbf{0.533} \\
\hline
InternLM2.5	& B+R+S	& -& 0.575	& 0.500	& \textbf{0.564}	& \textit{0.500}	& \textit{0.480}	& 0.333	& 0.492 \\
\hline
InternLM2.5	& B+SS	& -& \textit{0.583} & \textbf{0.536}	& \textit{0.500}	& \textbf{0.570}	& \textit{0.480}	& \textbf{0.450}	& \textit{0.520} \\
\hline \hline
InternLM2.5	& B	& Text& 0.517	& 0.445	& 0.463	& 0.530	& 0.390	& 0.467	& 0.469 \\
\hline
InternLM2.5	& B+R	& Text& 0.567	& 0.454	& \textit{0.545}	& 0.490	& 0.370	& 0.450	& 0.480 \\
\hline
InternLM2.5	& B+S	& Text& 0.616	& \textit{0.500}	& \textit{0.545}	& \textit{0.600}	& \textbf{0.550}	& 0.467	& 0.546 \\
\hline
InternLM2.5	& B+R+S	& Text& \textbf{0.683}	& 0.454	& \textbf{0.563}	& 0.570	& \textit{0.530}	& \textbf{0.500}	& \textit{0.550} \\
\hline
InternLM2.5	& B+SS	& Text& \textit{0.658}  & \textbf{0.581}	& \textbf{0.563}	& \textbf{0.640}	& 0.520	& \textit{0.483}	& \textbf{0.574} \\
\hline

\end{tabular}
\label{tab:ai_acc}
\end{table*}



Next, we investigated the ability of AI systems to reason about how humans might perform in the word-guessing task. In particular, we provided two AI systems, InternLM \cite{2023internlm} and InternVL \cite{chen2024internvl}, with some of the information we asked our participants to share in the form of text. We then asked these AI systems to predict how likely each participant is to correctly guess each target word. The two AI systems differ in whether they can receive images as inputs or not. For InternVL (VL for vision-language), we directly pass in the image together with the textual prompt including instructions for the prediction, the information related to the participant, the target sentence and target word. For InternLM (LM for language model), we investigated two alternatives to providing the image: one where the image is completely omitted, and the other where we instead pass a detailed description of the image, obtained from InternVL, then all reasoning is performed in language space only. 
In all cases, we provided the language background of the participant, as written in their form response, in the format \texttt{(Language1,Proficiency1), ..., (LanguageN,ProficiencyN)}. 
For Korean, we experimented with normal script and Roman (Latin) transliteration.

Given this information, we ask the system to provide a likelihood that the participant with this language background will guess the target word in the target sentence correctly (with 0\%, 25\%, 50\%, 75\% and 100\% as the options). If the system estimated a likelihood of 75\% or more and the person did indeed guess correctly, or if the system estimated likelihood of 50\% or less and the person guessed incorrectly, we gave the AI system a score of 1, otherwise a score of 0. This design is to simplify the scoring and make the comparison across languages cleaner, because the confidence score provided by AI systems may be inaccurate or biased. 

We tested AI systems' predictions of human performance. We varied the system (VL vs LM) and features: Background only (B), Background and a list of self-reported Recalled Words along with their translations (B+R), Background and the self-reported strategy the participant used to guess the word (B+S), and a combination of all three (B+R+S). 

In some cases, the strategy ``leaks'' some problem-specific information that oversimplifies the task for the AI system (e.g. ``I knew the word''). Thus we also experimented with a more generalized setting using a \emph{summary} of strategies participants use in general, without providing the exact strategy used for this particular image-text pair. To obtain the summary, we ask GPT-4o to summarize the strategies participants reported using across the ten examples and all languages, and do light post-processing of the output to merge redundant strategy categories and remove less sensible ones. We refer to this feature as Background and Strategy Summary (B+SS). The text of the summary is as follows:
``(1) Recognizing Cognates and Similar Words; Using Phonetic or Alphabetic Similarity. (A) Some users guessed based on how the word sounded when pronounced, comparing it to words in English or other languages.
(B) Others identified similar spelling patterns between the foreign word and English words.
(2) Using the Image for Context.
(A) Many users relied on the image to infer the meaning of the word.
(B) If they didn’t recognize any words in the sentence, they simply guessed the most prominent object in the image.
(3) Inferring from Context and Logical Deduction. Analyzing Sentence Structure and Word Position.
(A) Some users examined the position of the target word in the sentence to determine its function (e.g., if it followed ``une,'' it was likely a noun).
(B) Some users inferred the word’s meaning based on surrounding words and logical elimination.
(C) A few used grammatical reasoning, such as assuming adjectives described objects seen in the image.''

Table \ref{tab:ai_acc} shows the result in percentage, indicating what fraction of possible points each system-feature configuration received. Performance is low overall, and many results are close to chance (a random guess would give accuracy of 50\% on average if the data is balanced). Using the image is better than not using it, but using a text description and InternLM is generally better than directly using the image for reasoning with InternVL. The language where it is hardest to predict guessing performance is Turkish (based on performance in the best overall setting, InternLM-Text-B+SS), while the easiest one is Spanish. 

We observe that recalled words do not improve performance overall, compared to just using the background information. On the other hand, specific strategy information helps significantly when reasoning is done in language space only (LM), improving overall accuracy from 0.402 to 0.533 (i.e. by 32\%) when using no image information at all, and from 0.469 to 0.546 (16\% gain) when the image is described as text. Combining all three sources of information (B+R+S) does not significantly improve over just B+S. Adding strategy summary (B+SS) is better than using the specific example strategy (B+S) in the settings where the image is used, whether directly or described through text. 
We observe that the combination of language background and strategy summary has the best performance for both InternLM with image descriptions and InternVL; and second best performance for InternLM without image information. For InternVL which uses the image directly, B+SS improves accuracy by 19\% (from 0.476 to 0.568) comparing to B. These increases in performance exhibit that B+SS is a promising feature combination, opening up the possibility of real-world use of these features given a summary of relevant strategies. 
Looking at performance for individual languages under the InternLM setting using images described as text (bottom group),
B+S and B+SS gain over B in all languages except Turkish, which merits further exploration.

Finally, we also conducted an experiment where we put the AI system in the shoes of the human participants more directly: we replaced the target word with a blank token ``[BLANK]'', and asked InternVL to guess the meaning of the word. We considered the response to be correct if the predicted word was either a hyponym or hypernym of the correct (original) word. The performance of the AI system was lower than human performance overall except for Korean. In Korean, where characters are presented in the normal form, InternVL produced a correct response in 8 out of 10 cases. The number in Spanish was 6, in French and German 5, and in Turkish 4.
We also examined which of the ten examples were easier or harder for the AI system, and found this difficulty to be uncorrelated with difficulty for our human participants. The accuracy averaged across languages for each of the 10 examples based on the system's predictions 
had moderate (0.529) but insignificant ($p$ = 0.359) Spearman correlation with accuracy for the 10 examples based on human predictions. This indicates significant potential to improve AI system ability to reason about, anticipate or mimic human performance on this task.
\section{CONCLUSIONS}
\label{sec:conclusion}

We investigated the task of participants guessing the meaning of a word from multimodal context (image and a sentence containing the word). We analyzed features of the images and text, as well as the participants' language background, and examined correlations between these factors and word guessing success. We found that language background was a reasonable predictor of success for some languages, but its correlation with success was weak in others. Further, we only found one image and one text feature to be significantly correlated with guessing success. We hope our work inspires future work in analyzing what makes meaning inference in a multimodal context successful for learners of a foreign language. This could pave the way for designing learning opportunities in a multimodal setting that are both enjoyable, akin to a real-world experience, and appropriately challenging. We acknowledge some limitations: our work does not currently safeguard against AI biases, we use limited datapoints, and do not examine more free-form captions, such as those users might use on social media.

\addtolength{\textheight}{-12cm}   









\end{document}